\documentclass{article}

\usepackage{PRIMEarxiv}

\usepackage{enumitem}
\usepackage{amsmath}
\usepackage{mathtools}
\usepackage{amsthm}
\usepackage{longtable}
\usepackage{comment}
\usepackage{indentfirst}

\usepackage{natbib}

\usepackage[utf8]{inputenc}
\usepackage[T1]{fontenc}
\usepackage{hyperref}
\usepackage{url}
\usepackage{booktabs}
\usepackage{amsfonts}
\usepackage{amssymb}
\usepackage{nicefrac}
\usepackage{microtype}
\usepackage{graphicx}
\usepackage{multirow}
\usepackage{array}
\usepackage{siunitx}
\usepackage{xcolor}
\usepackage{listings}
\usepackage{graphics}
\usepackage{subcaption}
\usepackage{caption}

\definecolor{codegreen}{rgb}{0,0.6,0}
\definecolor{codegray}{rgb}{0.5,0.5,0.5}
\definecolor{codepurple}{rgb}{0.58,0,0.82}
\definecolor{backcolour}{rgb}{0.97,0.97,0.97}

\title{\includegraphics[width=\textwidth]{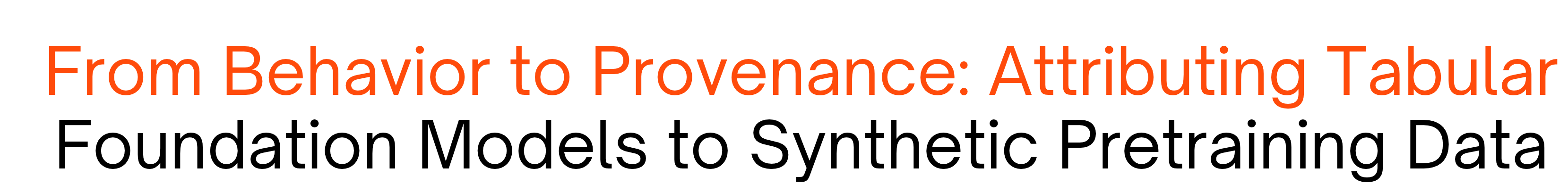}}

\author{
  Mohamed Bouadi$^*$, Nassim Bouarour$^*$,\\
  Shivam Dubey, Aditya Tanna,\\ 
  Vinay Kumar Sankarapu \\
  \affiliation{Lexsi Labs}\\
   \small{$^*$Equal contribution}
}

\setabstract{%
Training-data attribution aims to identify which training examples shape model behavior, yet validating such claims is difficult because causal training influence is rarely observable. We argue that controlled synthetic pretraining makes attribution experimentally testable. Using O'PRIOR, a provenance-rich synthetic task generator for tabular foundation models, we construct a testbed in which every pretraining task carries explicit lineage over structural mechanisms, missingness, confounding, shortcuts, and distribution shift. We combine behavior-conditioned attribution with counterfactual retraining and provenance-aware interventions to test both task-level faithfulness and mechanism-level consistency. On held-out real tasks, removing the top-attributed $5\%$ of synthetic tasks decreases mean ROC-AUC by $0.013$, compared with $0.002\pm0.004$ under random removal, while removing bottom-attributed tasks improves performance by $0.003$. Within shortcut-provenance tasks, targeted removal yields an effect of $0.043$ versus $0.016$ for matched random removal. Provenance discrimination is more modest by ranking AUROC ($0.55$--$0.62$), despite substantial top-$k$ enrichment, revealing that provenance association and interventional faithfulness need not coincide. Our results establish synthetic provenance as a controlled setting for verifiable contributive attribution
}

\setkeywords{Training Data Attribution; Synthetic Priors; Data Provenance; Tabular Foundation Models}

\runningtitle{From Behavior to Provenance: Attributing Tabular Foundation Models to Synthetic Pretraining Data}

\begin{document}
\maketitle

\section{Introduction}

Training-data attribution seeks to explain model behavior in terms of the data that shaped it. Given a prediction, capability, or failure, contributive attribution methods aim to identify the training examples that influenced that behavior. Existing approaches include influence functions, gradient-based tracing, data-models, and scalable estimators such as TRAK \citep{koh2017understanding,pruthi2020estimating,ilyas2022datamodels,park2023trak}. Yet a fundamental difficulty remains: \emph{how can we tell whether an attribution is faithful?} The contribution of a training example is not directly observed, and provenance metadata, when available, records where
data came from rather than how the learned model would have behaved had that data been absent. Attribution is therefore often evaluated through plausibility or indirect proxies rather than controlled interventions, limiting its use for auditing and accountability.

We argue that synthetic pretraining turns this validation problem into an experimental opportunity. When training data is produced by a controllable generator, each training task can carry an explicit record of how it was generated, while the training corpus itself can be manipulated. We study
this setting in tabular foundation models (TFMs) \citep{tabpfn-3.5,tabiclv2, orionmsp,orionbix,limix-2}, whose predictive behavior is largely acquired from synthetic supervised tasks. In particular, we build on O'PRIOR
\citep{bouadi2026shaping}, a compositional synthetic prior in which each task records its structural mechanism together with interventions for missingness, confounding, spurious correlation, and distribution shift. While O'PRIOR was originally introduced to study how prior design affects
downstream TFM quality, we use its controllable lineage for a different purpose: as a provenance-rich testbed for evaluating training-data attribution.

Crucially, known provenance is not itself causal ground truth. A task may carry a particular generative mechanism without being important for a given downstream behavior, and two tasks with identical provenance may have very different training effects. Our framework therefore connects three quantities that are normally unavailable together: the attribution assigned to individual synthetic tasks, the known provenance of those tasks, and the behavioral effect of removing them and retraining. At the \emph{task level}, we test interventional faithfulness by asking whether removing highly attributed tasks changes a downstream behavior more than removing equally sized random or weakly attributed subsets. At the \emph{provenance level}, we ask whether influential tasks concentrate on generative mechanisms associated with the behavior being explained, and whether this association changes coherently under controlled interventions. This separation lets us distinguish plausible provenance association from attribution that survives intervention on the training corpus.

Our experiments reveal a clear distinction between \emph{provenance association} and \emph{interventional faithfulness}. Influence is the most removal-faithful estimator: removing its top-attributed $5\%$ of tasks decreases mean ROC-AUC by $0.013$, versus $0.002\pm0.004$ under random
removal, while bottom removal slightly improves performance. Within shortcut provenance, Influence also yields the largest top-vs-random counterfactual gap ($0.030$). Yet provenance discrimination remains modest (macro ROC-AUC $0.55$--$0.62$), and TracIn achieves higher shortcut Lift@$5\%$ than
Influence ($1.81\times$ vs.\ $1.60\times$) despite weaker interventional faithfulness. Provenance-level attribution is also slightly more stable than task-level rankings ($0.86$ vs.\ $0.82$). Together, these results show that provenance association, behavioral influence under retraining, and
attribution stability are related but distinct properties.

\paragraph{Contributions.}
We make three contributions:
\begin{enumerate}
    \item \textbf{Provenance-grounded attribution:} we formulate synthetic-data attribution hierarchically, connecting downstream behavior to influential pretraining tasks and those tasks to their known generative provenance.
    \item \textbf{Interventional validation:} using O'PRIOR's frozen synthetic corpus, we test attribution through counterfactual task removal and provenance-conditioned interventions rather than attribution plausibility alone.
    \item \textbf{A systematic study of faithfulness, provenance association, and reliability:} we compare attribution methods against intervention-derived effects, study attribution under controlled failures and distribution shifts, and show that these evaluation axes need not rank attribution methods in the same way.
\end{enumerate}

\section{Provenance-Grounded Attribution}
\label{sec:framework}

Training-data attribution seeks to estimate how training examples contribute to learned model behavior, yet the causal quantity of interest is generally unobserved. Given a model $f_\theta$ trained on $\mathcal{D}$, an attribution method returns scores intended to identify which elements of $\mathcal{D}$ contributed to a behavior $B$. Whether these scores reflect genuine training influence is difficult to establish: the causal contribution of an example is unobserved, and dataset provenance, when available, does not reveal how model behavior would change had that example been absent. Our framework is built around removing that obstacle by keeping three quantities explicitly separate: the \emph{known generative provenance} $z_i$ of a synthetic training task, an \emph{attribution method's estimate} $a_i^{(B)}$ of its contribution to $B$, and the \emph{interventional effect} $\tau_B(S)$ of actually removing a set of tasks $S$ and retraining. Provenance is recorded lineage, not assumed causal ground truth; whether it agrees with $a^{(B)}$ and $\tau_B$ is the empirical question the rest of the paper answers.

\subsection{Setup and behavior-conditioned attribution}
\label{sec:setup}

\paragraph{A frozen, provenance-annotated corpus.} 
Let $\mathcal{D} = \{D_1, \ldots, D_N\}$ be a synthetic pretraining corpus of in-context supervised tasks $D_i = (X_i^s, y_i^s, X_i^q, y_i^q)$, instantiated once from O'PRIOR \citep{bouadi2026shaping} and frozen rather than resampled online, so that individual tasks remain identifiable across scoring and retraining. 

O'PRIOR's generation is compositional: structural mechanism, observational transformation, missingness, confounding, shortcut injection, and support–query shift are each controlled independently rather than folded into a single generator identity \cite{bouadiyou}. Each task therefore carries provenance $z_i = (z_i^{\text{str}}, z_i^{\text{obs}}, z_i^{\text{shift}}, \omega_i)$, recording which mechanisms were active and their severities, and the corpus defines a provenance matrix $Z = [z_1^\top, \ldots, z_N^\top]^\top$. 

Note that $z_i$ answers \emph{what process generated $D_i$}, not \emph{whether that process caused a downstream behavior}, the latter requires intervention and is what Section~\ref{sec:faithfulness} tests.

\paragraph{Behavior-conditioned attribution.}
We evaluate a trained model $\theta=\mathcal{A}(\mathcal{D};\xi)$ on held-out real tasks $\mathcal{R}=\{R_1,\ldots,R_M\}$ never used for pretraining, and condition attribution on a behavior functional $B(\theta;\mathcal{R})$ rather than attributing the model globally. For each synthetic task, $a_i^{(B)}=\operatorname{Attr}(D_i,B;\theta)$ estimates its contribution to behavior $B$. Our primary estimator is a first-order influence-function approximation \citep{koh2017understanding},
\[
a_i^{(B)}
=
- g_B(\theta^\star)^\top
\widehat H_{\theta^\star}^{-1}
g_i(\theta^\star),
\qquad
g_i(\theta^\star)=\nabla_\theta\ell(D_i;\theta^\star),
\quad
g_B(\theta^\star)=\nabla_\theta\mathcal{L}_B(\mathcal{R};\theta^\star),
\]
where $\widehat H_{\theta^\star}^{-1}$ denotes the inverse-Hessian approximation used in our implementation and $g_i$ is averaged within each task so scores are not confounded with task size. For behaviors whose reported metric is non-differentiable (e.g., ROC-AUC), $\mathcal{L}_B$ denotes a differentiable surrogate used only for attribution, while $B$ is retained for counterfactual evaluation. The framework is estimator-agnostic: we compare Influence against multi-checkpoint TracIn \citep{pruthi2020estimating}, TRAK \citep{park2023trak}, and a representation-similarity baseline. Throughout, $a^{(B)}$ is treated as a hypothesis about training influence, not as ground truth.

We orient attribution scores so that larger $a_i^{(B)}$ corresponds to a larger positive contribution to the reported behavior $B$; when the differentiable objective $\mathcal{L}_B$ is a loss, we reverse the corresponding loss-influence sign as needed.

\subsection{Interventional validation and provenance analysis}
\label{sec:faithfulness}

We validate that hypothesis by manipulating the corpus itself. For $S \subseteq \mathcal{D}$, define the counterfactually retrained model 
\[
\theta_{-S,\xi} = \mathcal{A}(\mathcal{D} \setminus S; \xi)
\]
and the removal effect 
\[
\tau_B(S) = \mathbb{E}_\xi\!\left[ B(\theta_{\mathcal{D},\xi}; \mathcal{R})
- B(\theta_{-S,\xi}; \mathcal{R}) \right]
\]
holding the architecture, optimizer, and evaluation protocol fixed across removal conditions, while matching random seeds across retraining runs. Thus, \(\tau_B(S)\) is the causal effect of this finite-corpus removal intervention under training algorithm \(\mathcal A\), rather than an intrinsic influence value independent of the learning procedure. For a removal fraction $r$, let $S_{\text{top}}^r$ and $S_{\text{bottom}}^r$ be the top- and bottom-$rN$ tasks under $a^{(B)}$, and $S_{\text{rand}}^r$ a size-matched random subset. We say $a^{(B)}$ exhibits \emph{removal faithfulness} if
\[
\tau_B(S_{\text{top}}^r) > \mathbb{E}_{S_{\text{rand}}^r}\!\left[\tau_B(S_{\text{rand}}^r)\right]
\]
and, ideally,
\[
\tau_B(S_{\text{top}}^r) > \tau_B(S_{\text{bottom}}^r).
\]
Gradient similarity or representational resemblance to a behavior is not by
itself evidence of interventional contribution; removal faithfulness asks whether the behavior actually changes once nominated tasks are unavailable to learning, giving an intervention-based reference against which any attribution method can be scored. 

\paragraph{From influential tasks to generative mechanisms.}
Synthetic provenance supports a second question unavailable to standard attribution benchmarks: which generative properties characterize the tasks that matter. For mechanism $c$ with indicator $z_{ic}$, we first measure whether $c$ is overrepresented among highly attributed tasks relative to its base rate in the corpus,
\[
\operatorname{Enrich}_c^{(B,r)} =
\frac{P(z_{ic}=1 \mid D_i \in S_{\text{top}}^r)}{P(z_{ic}=1)}.
\]
Because O'PRIOR mechanisms co-occur, enrichment alone cannot support a mechanism-level causal claim, so we pair it with a mechanism-conditioned intervention: within the stratum $\mathcal{D}_c = \{D_i : z_{ic}=1\}$, we compare removing the most highly attributed mechanism-$c$ tasks against removing an equal number of randomly chosen tasks from the same stratum. A gap between the two, $\tau_B(S_{\text{top},c}^r) > \mathbb{E}[\tau_B(S_{\text{rand},c}^r)]$, indicates that attribution distinguishes consequential instances \emph{within} a mechanism, not just membership in it, a property we call \emph{within-provenance faithfulness}.

\paragraph{Mechanism membership vs.\ ranking.}
We distinguish two complementary questions. A \emph{membership effect} asks whether removing tasks carrying mechanism $c$ changes behavior $B_c$ at all, whereas \emph{within-provenance faithfulness} asks whether attribution can identify the most consequential tasks \emph{among} those carrying $c$. The former tests whether exposure to a provenance class is behaviorally relevant; the latter tests whether the attribution ranking contains additional information beyond provenance membership. We report both quantities because provenance association alone need not imply faithful ranking within the corresponding mechanism.

\paragraph{Behavior specificity.}
Attribution should depend on the behavior being explained. For behaviors $\mathcal{B}=\{B_{\mathrm{iid}},B_{\mathrm{miss}},B_{\mathrm{shift}}, B_{\mathrm{shortcut}},\ldots\}$, we stack their task-attribution vectors into $A\in\mathbb{R}^{N\times|\mathcal{B}|}$ and study their alignment with the known provenance matrix $Z$. We quantify this alignment through base-rate-normalized top-$k$ enrichment and, for graded interventions, the Spearman correlation $\rho_{\mathrm{sev}}$ between intervention severity and attribution to the behavior-matched provenance. Thus, the question is not whether attribution recovers a single predetermined
mechanism---O'PRIOR mechanisms are compositional---but whether the distribution
of attributed provenance changes systematically with the behavior being explained.

\subsection{Reliability}
\label{sec:reliability}

Individual synthetic tasks can be redundant or mutually substitutable, so task-level attribution may be unstable even when the underlying training signal is not: two runs may rely on different examples while learning similar behavior. We therefore ask whether provenance-level attribution is more reproducible. For attribution vector $a^{(s,B)}$ obtained under seed or perturbation $s$, define the signed normalized scores 
\[
\tilde a_i^{(s,B)} = \frac{a_i^{(s,B)}}{\sum_j |a_j^{(s,B)}|+\epsilon},
\]  
and the prevalence-normalized provenance projection
\[
p_c^{(s,B)} = \frac{N}{\sum_i z_{ic}} \sum_i \tilde a_i^{(s,B)}z_{ic}.
\]

We compare task-level stability 
\[
\mathcal{S}_{\mathrm{task}} = \mathbb{E}_{s\neq s'}[\rho(a^{(s,B)},a^{(s',B)})]
\]
with provenance-level stability
\[
\mathcal{S}_{\mathrm{prov}} = \mathbb{E}_{s\neq s'}[\rho(p^{(s,B)},p^{(s',B)})]
\]
where $\rho$ is Spearman rank correlation. This comparison tests whether aggregating attribution to generative mechanisms preserves a more reproducible signal than individual-task rankings.

Together, the framework makes three quantities jointly accessible that are usually separated in training-data attribution: estimated task attribution $a^{(B)}$, observed behavioral change under counterfactual removal $\tau_B$, and known generative provenance $Z$. Their combination lets us distinguish three properties that need not coincide: \emph{removal faithfulness}, whether
attributed training tasks actually affect the behavior they are credited with; \emph{provenance association}, whether highly attributed tasks are enriched for behavior-relevant generative mechanisms; and \emph{attribution reliability}, whether these conclusions persist across runs and perturbations. The contribution is therefore not a new attribution estimator, but a controlled evaluation paradigm in which apparent provenance alignment can be tested against intervention-derived training effects.

\section{Experiments}
\label{sec:experiments}

\subsection{Experimental protocol}
\label{sec:protocol}

\paragraph{Frozen synthetic corpus.}
We instantiate O'PRIOR once to obtain a fixed corpus $\mathcal{D}=\{D_i\}_{i=1}^{N}$ of $N=5{,}000$ synthetic classification tasks. Each task contains an in-context support/query split, a persistent identifier, and provenance $z_i$ recording its structural mechanism and active missingness, confounding, shortcut, and distribution-shift interventions. Crucially, the corpus is frozen before training: no new synthetic tasks are sampled during either pretraining or counterfactual retraining, and a removed task remains unavailable throughout the corresponding retraining run. Individual synthetic tasks therefore remain well-defined units of both attribution and intervention.

\paragraph{Model and downstream behaviors.}
We train NanoTabPFN \citep{nanotabpfn,tabpfnv2} for $15$ epochs with $2{,}000$ optimization updates per epoch and four synthetic tasks per batch. Evaluation uses $22$ held-out OpenML-CC18 classification datasets \citep{openmlcc18}; no real dataset is used during synthetic pretraining. The default behavior $B_{\mathrm{iid}}$ is mean ROC-AUC across these real tasks. We additionally construct controlled missingness, covariate-shift, and shortcut behaviors to test whether attribution changes with the behavior being explained. When the reported behavior is non-differentiable, we use cross-entropy only as the differentiable attribution objective while retaining ROC-AUC for behavioral evaluation.

\paragraph{Attribution and counterfactual controls.}
Influence is our primary attribution estimator, with TracIn \citep{pruthi2020estimating}, TRAK \citep{park2023trak}, and representation similarity as comparisons. For counterfactual validation, models are retrained from scratch with the same architecture and number of optimization updates as the full-corpus model; only the available pretraining tasks differ. This controls for training-budget effects and makes the intervention the removal of a specified subset of the fixed corpus. We report $\Delta B=B_{\mathrm{full}}-B_{\mathrm{removed}}$, so positive values indicate that removing the selected tasks degrades downstream performance. Random-removal controls use four independently sampled subsets with matched training runs and are reported as mean $\pm$ empirical standard deviation.

\begin{table*}[t]
\centering
\small
\caption{
\textbf{Attribution quality on the frozen O'PRIOR corpus} ($N{=}$5k). Global removal quantities use $B_{\mathrm{iid}}$ at $r=10\%$; Lift@$5\%$ is base-rate-normalized matched-provenance enrichment, and Within-prov.\ is the top-vs-random gap restricted to shortcut tasks. $\mathcal{S}$ denotes mean pairwise Spearman stability. Higher is better.
}
\label{tab:mainresults}
\begin{tabular}{lcccccc}
\toprule
Method &
$\Delta_{\mathrm{top}}$ &
Faith.\ gap &
Lift@$5\%$ (matched) &
Within-prov.\ gap &
$\mathcal{S}_{\mathrm{task}}$ &
$\mathcal{S}_{\mathrm{prov}}$ \\
\midrule
Influence & $+0.012$ & $+0.008$ & $1.60\times$ & $+0.030$ & $0.85$ & $0.91$ \\
TracIn    & $+0.005$ & $+0.002$ & $1.81\times$ & $+0.023$ & $0.80$ & $0.84$ \\
TRAK      & $+0.004$ & $+0.001$ & $1.10\times$ & $+0.008$ & $0.78$ & $0.84$ \\
Rep.\ sim.& $-0.002$ & $-0.005$ & $1.00\times$ & $-0.002$ & $0.70$ & $0.72$ \\
\bottomrule
\end{tabular}
\end{table*}


\begin{figure*}[t]
    \begin{subfigure}[t]{0.32\textwidth}
        \centering
        \includegraphics[width=\linewidth]{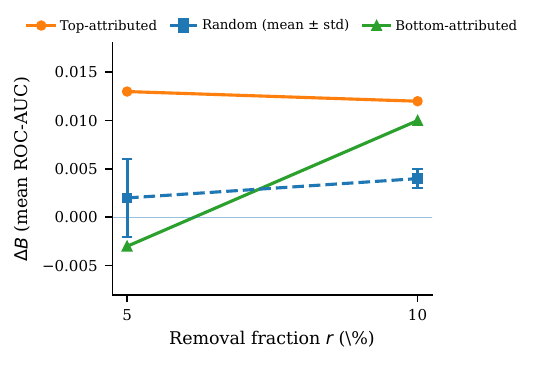}
        \caption{Global removal under Influence.}
        \label{fig:faithfulness-a}
    \end{subfigure}
    \hfill
    \begin{subfigure}[t]{0.32\textwidth}
        \centering
        \includegraphics[width=\linewidth]{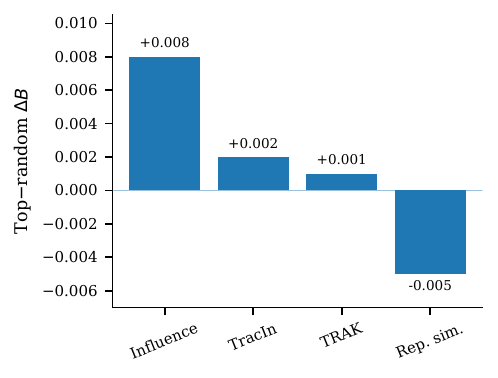}
        \caption{IID top-vs-random gap.}
        \label{fig:faithfulness-b}
    \end{subfigure}
    \hfill
    \begin{subfigure}[t]{0.32\textwidth}
        \centering
        \includegraphics[width=\linewidth]{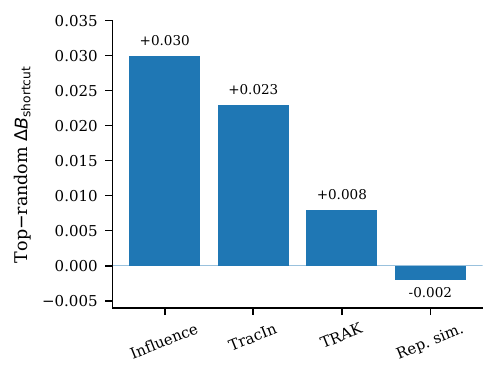}
        \caption{Within-shortcut gap.}
        \label{fig:faithfulness-c}
    \end{subfigure}
    \caption{\textbf{Counterfactual faithfulness of training-data attribution}. \textbf{(a)} Change in mean ROC-AUC after removing top-attributed, random, or bottom-attributed tasks; random removal shows mean
    $\pm$ empirical standard deviation across four matched runs.
    \textbf{(b)} Top-vs-random removal gap for $B_{\mathrm{iid}}$ at
    $r=10\%$ across attribution estimators.
    \textbf{(c)} Within-provenance faithfulness under
    $B_{\mathrm{shortcut}}$, comparing top-attributed and random removal restricted to shortcut-generated tasks. Positive gaps indicate that attribution identifies subsets whose removal is more consequential than the matched random control.
    }
    \label{fig:faithfulness}
\end{figure*}

\subsection{Task-level faithfulness and provenance association}
\label{sec:exp_faithfulness}

\paragraph{E1: Task-level removal faithfulness.}
We first test the most direct claim made by contributive attribution:
whether tasks assigned high influence exert larger behavioral effects when they are made unavailable to learning. We rank all $5{,}000$ synthetic tasks with respect to $B_{\mathrm{iid}}$ and retrain after removing top-attributed, bottom-attributed, or size-matched random subsets.

At $r=5\%$, removing the top-attributed tasks decreases mean ROC-AUC by
$\Delta B=0.013$, compared with $0.002\pm0.004$ under random removal
(Fig.~\ref{fig:faithfulness-a}). Removing the bottom-attributed subset
instead gives $\Delta B=-0.003$, i.e., a small improvement, yielding a
top-vs-random faithfulness gap of $0.011$. The contrast is therefore visible against both controls: the highly attributed subset is substantially more consequential than random removal, whereas the lowest-ranked tasks are not required to preserve the baseline behavior.

The effect persists at $r=10\%$: top removal gives $\Delta B=0.012$ versus $0.004$ for random removal, corresponding to a gap of $0.008$. Separation from the bottom subset is weaker at this larger intervention, so we regard the $5\%$ experiment as the cleaner test of
fine-grained ranking faithfulness and the $10\%$ intervention as evidence that the top-vs-random effect is not confined to a single removal fraction. Importantly, these results validate the attribution ranking through the learning procedure itself rather than through similarity to downstream examples or provenance labels.

The estimator comparison reinforces this distinction
(Fig.~\ref{fig:faithfulness-b}). At $r=10\%$, Influence yields the
largest top-vs-random gap ($+0.008$), compared with $+0.002$ for TracIn
and $+0.001$ for TRAK; representation similarity produces a negative gap
($-0.005$). Thus, methods that generate apparently meaningful rankings
need not identify subsets whose removal has a correspondingly large
behavioral effect. Under this intervention, Influence provides the clearest
task-level faithfulness signal.


\paragraph{E2: Provenance association and within-provenance faithfulness.}
Counterfactual removal establishes that some attributed tasks matter, but does not explain what properties characterize them. Because every O'PRIOR task carries explicit generative provenance, we can additionally ask whether high attribution concentrates on mechanisms related to the behavior being explained. We evaluate this in two complementary ways: global ranking discrimination and base-rate-normalized top-$k$ enrichment.

Attribution contains statistically detectable, but far from complete, information about generative provenance. Behavior-matched provenance discrimination is approximately $0.60$ ROC-AUC, with macro values ranging from $0.55$ to $0.62$ across provenance columns (Fig.~\ref{fig:provenance-b}); matched average precision is likewise above the corresponding base rate. These values are meaningfully above chance but modest in absolute terms, so we do not interpret the attribution score as recovering the generative mechanism from the full task ranking.

The head of the ranking shows a clearer signal. Behavior-matched Lift@$5\%$ is approximately $1.3\times$ overall, and shortcut provenance reaches $1.60\times$ under Influence for $B_{\mathrm{shortcut}}$. Matched provenance scores are also shifted relative to their controls (Wilcoxon $z>2.23$). The combination of moderate global ROC-AUC and stronger top-$k$ enrichment indicates that provenance information is concentrated most clearly among the highest-attributed tasks rather than being uniformly separable throughout the full ranking.

Provenance association alone, however, does not establish that the ranking distinguishes consequential tasks \emph{within} a provenance class. We therefore restrict the intervention to shortcut-generated tasks and compare the highest-attributed subset with a size-matched random subset from the same stratum. Influence yields a within-shortcut top-vs-random gap of
$+0.030$, followed by TracIn ($+0.023$), TRAK ($+0.008$), and representation similarity ($-0.002$)
(Fig.~\ref{fig:faithfulness-c}). Because mechanism membership is held fixed in this comparison, the positive gap cannot be explained merely by the presence of shortcut provenance: the attribution ranking carries additional information about which shortcut-generated tasks are behaviorally
consequential.

The result is not universal across all behavior--mechanism combinations. Other targeted pairs show heterogeneous or inconclusive separation. We therefore treat within-provenance faithfulness as an empirical property to be tested for each behavior rather than assuming that known provenance is itself causal ground truth.


\subsection{Behavior specificity and estimator disagreement}
\label{sec:exp_behavior}

\begin{figure*}[t]
    \centering

    \begin{subfigure}[t]{0.46\textwidth}
        \centering
        \includegraphics[width=\linewidth]{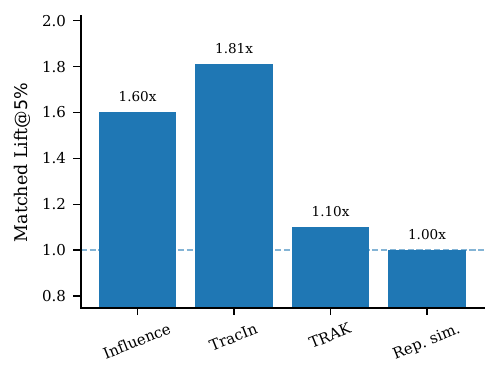}
        \caption{Behavior-matched provenance enrichment.}
        \label{fig:provenance-a}
    \end{subfigure}
    \hfill
    \begin{subfigure}[t]{0.46\textwidth}
        \centering
        \includegraphics[width=\linewidth]{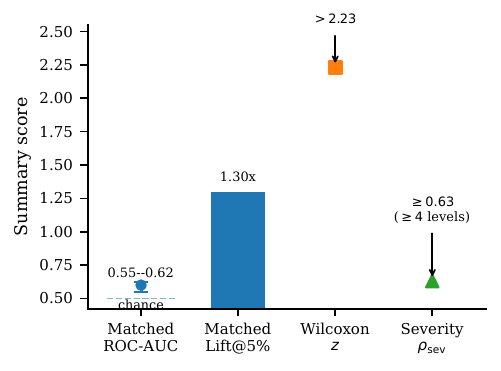}
        \caption{Provenance and behavior-specificity evidence.}
        \label{fig:provenance-b}
    \end{subfigure}

    \caption{
    \textbf{Provenance association and behavior specificity.}
    \textbf{(a)} Lift@$5\%$ of behavior-matched provenance among
    highly attributed tasks across estimators; $1\times$ denotes the
    corpus base rate. For the shortcut behavior, TracIn attains higher
    enrichment than Influence despite weaker counterfactual faithfulness.
    \textbf{(b)} Complementary provenance evidence: matched provenance
    discrimination is approximately $0.60$ ROC-AUC (macro range
    $0.55$--$0.62$), matched Lift@$5\%$ is approximately $1.3\times$
    overall, matched comparisons yield Wilcoxon $z>2.23$, and attribution
    to behavior-matched provenance tracks intervention severity with
    $\rho_{\mathrm{sev}}\ge0.63$ over at least four severity levels.
    Statistics in (b) have different interpretations and are not intended
    for direct comparison by magnitude.
    }
    \label{fig:provenance}
\end{figure*}

\paragraph{E3: Attribution tracks the behavior being explained.}
A behavior-conditioned attribution should not be invariant to a controlled change in the downstream behavior. We therefore keep the pretrained model and synthetic corpus fixed while varying the severity of missingness, covariate shift, and shortcut interventions at evaluation time and recompute attribution for each resulting behavior.

Across these graded interventions, attribution to behavior-matched provenance changes systematically with intervention severity. The corresponding Spearman correlations satisfy $\rho_{\mathrm{sev}}\ge0.63$ over at least four distinct severity levels (Fig.~\ref{fig:provenance-b}). Thus, the provenance associated with highly attributed training tasks changes monotonically with the strength of the downstream failure being explained rather than remaining fixed across evaluation conditions.

The response is heterogeneous across mechanisms: some controlled behaviors produce clear matched-provenance shifts whereas others are weaker or indistinguishable from controls. This negative evidence is important. Known provenance makes such failures visible and prevents behavior-specific attribution from being judged only by examples for which the expected mechanism happens to appear.


\paragraph{E4: Provenance association and interventional faithfulness can
disagree.}
Comparing estimators exposes a distinction that is difficult to observe in standard attribution benchmarks. Influence gives the largest global top-vs-random removal gap and the largest within shortcut counterfactual gap (Table~\ref{tab:mainresults}). Provenance enrichment gives a different
ordering: under $B_{\mathrm{shortcut}}$, TracIn achieves Lift@$5\%$ of $1.81\times$, exceeding Influence at $1.60\times$ (Fig.~\ref{fig:provenance-a}), even though its within-provenance faithfulness gap is smaller ($+0.023$ vs.\ $+0.030$).

This reversal shows that the two evaluations measure genuinely different properties. A method can preferentially retrieve tasks carrying the \textit{right} provenance label without identifying the subset whose absence has the largest effect on learning. Conversely, intervention-derived
faithfulness does not require perfect recovery of the corresponding generative mechanism. We therefore treat provenance association as supporting evidence about \emph{where} attribution concentrates and counterfactual retraining as evidence about whether the nominated training data is behaviorally consequential under the specified learning intervention.


\subsection{Attribution reliability}
\label{sec:exp_reliability}

\begin{figure}[t]
    \centering
    \includegraphics[width=0.68\linewidth]{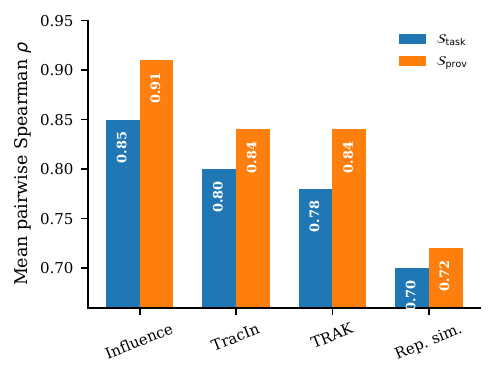}
    \caption{
    \textbf{Attribution reliability.}
    Mean pairwise Spearman stability at the individual-task and
    provenance-aggregated levels. Provenance aggregation improves
    stability for every estimator, although the gain is modest.
    }
    \label{fig:stability}
\end{figure}

\paragraph{E5: Task-level versus provenance-level stability.}
Finally, we ask whether attribution conclusions are reproducible across seed and checkpoint variants. At the level of individual training tasks, small changes in estimation can alter the ordering of examples that are partially redundant or substitutable. We therefore compare task-level rankings with attribution aggregated onto the recorded generative mechanisms.

Figure~\ref{fig:stability} shows that provenance-level summaries are more stable for every estimator considered. Influence increases from $\mathcal{S}_{\mathrm{task}}=0.85$ to $\mathcal{S}_{\mathrm{prov}}=0.91$; TracIn from $0.80$ to $0.84$; TRAK from $0.78$ to $0.84$; and representation similarity from $0.70$ to $0.72$. The improvement is therefore systematic but modest rather than dramatic. This supports provenance aggregation as a slightly more reproducible level of analysis without implying that it eliminates estimator uncertainty.

We distinguish this notion of stability from cross-estimator agreement. Influence and TracIn, for example, can each be internally reproducible while still disagreeing about which tasks should receive high attribution because they encode different notions of training influence. Reliability under repeated estimation and agreement across attribution assumptions are therefore separate properties.

\section{Conclusion}

We introduced a provenance-grounded framework for evaluating contributive training-data attribution in synthetically pretrained foundation models. The key idea is to keep three quantities explicitly separate: the attribution assigned to individual training tasks, their recorded generative provenance,
and the behavioral effect of removing those tasks and retraining. This makes it possible to test attribution claims through intervention rather than treating either attribution scores or provenance labels as ground truth.

Across our experiments, these quantities provide complementary but distinct signals. Influence exhibits the strongest counterfactual removal faithfulness, both globally and within shortcut-provenance strata, while attribution scores recover generative provenance only moderately at the level of the full ranking. At the same time, highly ranked tasks are enriched for behavior-matched provenance, and this attribution changes systematically under controlled failures and increasing intervention severity. Crucially, the methods need not be ordered consistently across these criteria: TracIn shows stronger shortcut-provenance enrichment than Influence ($1.81\times$ vs.\ $1.60\times$) despite a smaller within-provenance counterfactual gap ($0.023$ vs.\ $0.030$). This reversal demonstrates that \emph{provenance association and interventional faithfulness are genuinely
different properties}. We additionally find that aggregating task-level attribution onto generative mechanisms yields modestly more reproducible summaries, suggesting that provenance can provide a useful intermediate level of analysis between individual training examples and downstream behavior.

Taken together, these results suggest a broader role for controlled synthetic pretraining in attribution research. Known synthetic provenance should not be viewed as causal ground truth: it records how a task was generated, not whether that task caused a particular learned behavior. Its value instead comes from providing a structured reference that can be combined with counterfactual retraining, behavior-specific evaluation, and reliability analysis. In this sense, synthetic provenance turns attribution from a primarily plausibility-based exercise into one for which competing claims can be experimentally tested.

\paragraph{Limitations.}
Our study is deliberately controlled, and several limitations bound the scope of its conclusions. First, the experiments use a finite O'PRIOR corpus of $5{,}000$ synthetic tasks and a NanoTabPFN-scale tabular foundation model; the same relationships may change with substantially larger models, larger synthetic corpora, or natural-data pretraining. Second, the removal effect $\tau_B(S)$ is causal only with respect to the specific finite-corpus intervention and training algorithm considered here: it depends on the optimizer, training budget, initialization, and retraining procedure, and
should not be interpreted as an intrinsic influence value independent of learning dynamics. Third, removing sets of tasks can induce interaction and redundancy effects, so the effect of a subset need not equal the sum of the effects of its members. Fourth, O'PRIOR mechanisms are compositional and can co-occur; consequently, provenance enrichment provides evidence of association but does not isolate a single generative mechanism as the unique cause of a behavior. Our within-provenance controls mitigate this issue but do not remove all dependencies between mechanisms. Fifth, attribution for non-differentiable behaviors relies on a differentiable surrogate objective, which may not perfectly align with the reported downstream metric. Finally, counterfactual retraining is computationally expensive, which limits the number and scale of interventions that can be evaluated exhaustively.

Future work should extend this evaluation paradigm to larger foundation models, richer and multi-generation provenance graphs, and settings where synthetic and natural pretraining data coexist. An important direction is to develop cheaper validation procedures that retain the evidential value of
retraining-based interventions while scaling to substantially larger training corpora.

\bibliographystyle{unsrt}
\bibliography{references}

@article{orionmsp,
  author       = {Mohamed Bouadi and
                  Pratinav Seth and
                  Aditya Tanna and
                  Vinay Kumar Sankarapu},
  title        = {Orion-MSP: Multi-Scale Sparse Attention for Tabular In-Context Learning},
  journal      = {CoRR},
  volume       = {abs/2511.02818},
  year         = {2025},
  url          = {https://doi.org/10.48550/arXiv.2511.02818},
  doi          = {10.48550/ARXIV.2511.02818},
  eprinttype   = {arXiv},
  eprint       = {2511.02818},
  bibsource    = {dblp computer science bibliography, https://dblp.org}
}

@article{tabiclv2,
  author       = {Jingang Qu and
                  David Holzm{\"{u}}ller and
                  Ga{\"{e}}l Varoquaux and
                  Marine Le Morvan},
  title        = {TabICLv2: {A} better, faster, scalable, and open tabular foundation
                  model},
  journal      = {CoRR},
  volume       = {abs/2602.11139},
  year         = {2026},
  url          = {https://doi.org/10.48550/arXiv.2602.11139},
  doi          = {10.48550/ARXIV.2602.11139},
  eprinttype   = {arXiv},
  eprint       = {2602.11139},
  bibsource    = {dblp computer science bibliography, https://dblp.org}
}

@article{tabpfnv2,
  title={Accurate predictions on small data with a tabular foundation model},
  author={Hollmann, Noah and M{\"u}ller, Samuel and Purucker, Lennart and Krishnakumar, Arjun and K{\"o}rfer, Max and Hoo, Shi Bin and Schirrmeister, Robin Tibor and Hutter, Frank},
  journal={Nature},
  volume={637},
  number={8045},
  pages={319--326},
  year={2025},
  publisher={Nature Publishing Group UK London}
}

@article{openmlcc18,
  title={Openml benchmarking suites},
  author={Bischl, Bernd and Casalicchio, Giuseppe and Feurer, Matthias and Gijsbers, Pieter and Hutter, Frank and Lang, Michel and Mantovani, Rafael G and van Rijn, Jan N and Vanschoren, Joaquin},
  journal={arXiv preprint arXiv:1708.03731},
  year={2017}
}

@article{nanotabpfn,
  title={nanotabpfn: A lightweight and educational reimplementation of tabpfn},
  author={Pfefferle, Alexander and Hog, Johannes and Purucker, Lennart and Hutter, Frank},
  journal={arXiv preprint arXiv:2511.03634},
  year={2025}
}

@inproceedings{koh2017understanding,
  title={Understanding black-box predictions via influence functions},
  author={Koh, Pang Wei and Liang, Percy},
  booktitle={International conference on machine learning},
  pages={1885--1894},
  year={2017},
  organization={Pmlr}
}

@article{pruthi2020estimating,
  title={Estimating training data influence by tracing gradient descent},
  author={Pruthi, Garima and Liu, Frederick and Kale, Satyen and Sundararajan, Mukund},
  journal={Advances in neural information processing systems},
  volume={33},
  pages={19920--19930},
  year={2020}
}

@inproceedings{ilyas2022datamodels,
  title={Datamodels: Understanding predictions with data and data with predictions},
  author={Ilyas, Andrew and Park, Sung Min and Engstrom, Logan and Leclerc, Guillaume and Madry, Aleksander},
  booktitle={International Conference on Machine Learning},
  pages={9525--9587},
  year={2022},
  organization={PMLR}
}

@article{park2023trak,
  title={Trak: Attributing model behavior at scale},
  author={Park, Sung Min and Georgiev, Kristian and Ilyas, Andrew and Leclerc, Guillaume and Madry, Aleksander},
  journal={arXiv preprint arXiv:2303.14186},
  year={2023}
}

@article{bouadi2026shaping,
  title={Shaping the Prior: How Synthetic Task Distributions Determine Tabular Foundation Model Quality},
  author={Bouadi, Mohamed and Bouarour, Nassim and Kulkarni, Varun and Dubey, Shivam and Tanna, Aditya and Sankarapu, Vinay Kumar},
  journal={arXiv preprint arXiv:2605.18971},
  year={2026}
}

@inproceedings{bouadiyou,
  title={What You Pretrain On Matters: Synthetic Task Distributions Determine Tabular Foundation Model Quality},
  author={Bouadi, Mohamed and Bouarour, Nassim and Dubey, Shivam and Kulkarni, Varun and Tanna, Aditya and kumar Sankarapu, Vinay},
  booktitle={2nd ICML Workshop on Foundation Models for Structured Data}
}

@inproceedings{orionbix,
  title={Orion-Bix: Bi-axial attention for tabular in-context learning},
  author={Bouadi, Mohamed and Seth, Pratinav and Tanna, Aditya and Sankarapu, Vinay Kumar},
  booktitle={Proceedings of the ACM Web Conference 2026},
  pages={8673--8676},
  year={2026}
}

@article{tabpfn-3.5,
  title={TabPFN-3.5: Technical Report},
  author={J{\"a}ger, Benjamin and Erickson, Nick and Grinsztajn, L{\'e}o and Birkel, Felix and Fl{\"o}ge, Klemens and Key, Oscar and Kaya, K{\"u}r{\c{s}}at and K{\"u}bler, Jonas and Frankel, Ad{\`e}le and Schr{\"o}der, Tobias and others},
  journal={arXiv preprint arXiv:2609.17895},
  year={2026}
}

@article{limix-2,
  title={LimiX-2: A Contextual Mechanism Network Towards General Structured-Data Intelligence},
  author={Zhang, Xingxuan and Ren, Gang and Yuan, Hao and Zou, Hao and Tan, Hongze and Wang, Hui and Song, Jianhao and Li, Jiansheng and Zhang, Jiayao and Zhang, Jinghan and others},
  journal={arXiv preprint arXiv:2609.17488},
  year={2026}
}

\end{document}